\documentclass[12pt]{article}
\usepackage{scicite}
\usepackage{times}
\usepackage{amsmath}
\usepackage{amsfonts}
\usepackage{graphicx}
\usepackage[usenames, dvipsnames]{color}
\usepackage{multirow}
\usepackage{multibib}
\usepackage{tabularx, booktabs}
\usepackage{longtable}
\usepackage{caption}
\usepackage{colortbl}
\usepackage{siunitx}
\usepackage{url}
\usepackage{hyperref}
\usepackage{skak} 

\usepackage[top=1in, bottom=1in, left=1in, right=1in]{geometry}

\definecolor{darkgreen}{RGB}{0, 125, 0}
\definecolor{orange}{RGB}{255, 125, 125}
\definecolor{darkgray}{RGB}{64, 64, 64}
\newcolumntype{S}{>{\hsize=.3\hsize}X}

\title{Mastering Chess and Shogi by Self-Play with a General Reinforcement Learning Algorithm}
\date{}

\author
{
David Silver,$^{1\ast}$ 
Thomas Hubert,$^{1\ast}$
Julian Schrittwieser,$^{1\ast}$\\
Ioannis Antonoglou,$^{1}$
Matthew Lai,$^{1}$
Arthur Guez,$^{1}$
Marc Lanctot,$^{1}$\\
Laurent Sifre,$^{1}$
Dharshan Kumaran,$^{1}$
Thore Graepel,$^{1}$\\
Timothy Lillicrap,$^{1}$
Karen Simonyan,$^{1}$
Demis Hassabis$^{1}$
\\
\\
\normalsize{$^{1}$DeepMind, 6 Pancras Square, London N1C 4AG.}
\\
\normalsize{$^\ast$These authors contributed equally to this work.}
}

\begin{document}
\maketitle

\begin{abstract}
The game of chess is the most widely-studied domain in the history of artificial intelligence. The strongest programs are based on a combination of sophisticated search techniques, domain-specific adaptations, and handcrafted evaluation functions
that have been refined by human experts over several decades. In contrast, the \emph{AlphaGo Zero} program recently achieved superhuman performance in the game of Go, by \emph{tabula rasa} reinforcement learning from games of self-play. In this paper, we generalise this approach into a single \emph{AlphaZero} algorithm that  can achieve, \emph{tabula rasa}, superhuman performance in many challenging domains. Starting from random play, and given no domain knowledge except the game rules, \emph{AlphaZero} achieved within 24 hours a superhuman level of play in the games of chess and shogi (Japanese chess) as well as Go, and convincingly defeated a world-champion program in each case. 
\end{abstract}


The study of computer chess is as old as computer science itself. Babbage, Turing, Shannon, and von Neumann devised hardware, algorithms and theory to analyse and play the game of chess. Chess subsequently became the grand challenge task for a generation of artificial intelligence researchers, culminating in high-performance computer chess programs that perform at superhuman level \cite{Campbell02DeepBlue,FengHsiung02DeepBlue}. However, these systems are highly tuned to their domain, and cannot be generalised to other problems without significant human effort. 

A long-standing ambition of artificial intelligence has been to create programs that can instead learn for themselves from first principles \cite{samuel:recent}. Recently, the \emph{AlphaGo Zero} algorithm achieved superhuman performance in the game of Go, by representing Go knowledge using deep convolutional neural networks~\cite{maddison:deepgo,Silver16AG}, trained solely by reinforcement learning from games of self-play~\cite{Silver17AG0}. In this paper, we apply a similar but fully generic algorithm, which we call \emph{AlphaZero}, to the games of chess and shogi as well as Go, without any additional domain knowledge except the rules of the game, demonstrating that a general-purpose reinforcement learning algorithm can achieve, \emph{tabula rasa}, superhuman performance across many challenging domains.

 
A landmark for artificial intelligence was achieved in 1997 when Deep Blue defeated the human world champion \cite{Campbell02DeepBlue}. Computer chess programs continued to progress steadily beyond human level in the following two decades. These programs evaluate positions using features handcrafted by human grandmasters and carefully tuned weights, combined with a high-performance alpha-beta search that expands a vast search tree using a large number of clever heuristics and domain-specific adaptations. In the Methods we describe these augmentations, focusing on the 2016 Top Chess Engine Championship (TCEC) world-champion \emph{Stockfish}~\cite{Stockfish}; other strong chess programs, including Deep Blue, use very similar architectures~\cite{Campbell02DeepBlue,LevyNewborn09}.


Shogi is a significantly harder game, in terms of computational complexity, than chess~\cite{Allis94,Iida02Shogi}: it is played on a larger board, and any captured opponent piece changes sides and may subsequently be dropped anywhere on the board. The strongest shogi programs, such as Computer Shogi Association (CSA) world-champion \emph{Elmo}, have only recently defeated human champions \cite{WCSC27}. These programs use a similar algorithm to computer chess programs, again based on a highly optimised alpha-beta search engine with many domain-specific adaptations.


Go is well suited to the neural network architecture used in \emph{AlphaGo} because the rules of the game are translationally invariant (matching the weight sharing structure of convolutional networks), are defined in terms of \emph{liberties} corresponding to the adjacencies between points on the board (matching the local structure of convolutional networks), and are rotationally and reflectionally symmetric (allowing for data augmentation and ensembling). Furthermore, the action space is simple (a stone may be placed at each possible location), and the game outcomes are restricted to binary wins or losses, both of which may help neural network training. 

Chess and shogi are, arguably, less innately suited to \emph{AlphaGo}'s neural network architectures. The rules are position-dependent (e.g. pawns may move two steps forward from the second rank and promote on the eighth rank) and asymmetric (e.g. pawns only move forward, and castling is different on kingside and queenside). The rules include long-range interactions (e.g. the queen may traverse the board in one move, or checkmate the king from the far side of the board). The action space for chess includes all legal destinations for all of the players' pieces on the board; shogi also allows captured pieces to be placed back on the board. Both chess and shogi may result in draws in addition to wins and losses; indeed it is believed that the optimal solution to chess is a draw~\cite{Steinitz90Modern,Lasker65Common,Knudsen00Essential}.


The \emph{AlphaZero} algorithm is a more generic version of the \emph{AlphaGo Zero} algorithm that was first introduced in the context of Go \cite{Silver17AG0}. It replaces the handcrafted knowledge and domain-specific augmentations used in traditional game-playing programs with deep neural networks and a \emph{tabula rasa} reinforcement learning algorithm. 

Instead of a handcrafted evaluation function and move ordering heuristics, \emph{AlphaZero} utilises a deep neural network $(\mathbf{p},v) = f_\theta(s)$ with parameters $\theta$. This neural network takes the board position $s$ as an input and outputs a vector of move probabilities $\mathbf{p}$ with components $p_a = Pr(a|s)$ for each action $a$, and a scalar value $v$ estimating the expected outcome $z$ from position $s$, $v \approx \mathbb{E}[z | s]$. \emph{AlphaZero} learns these move probabilities and value estimates entirely from self-play; these are then used to guide its search. 

Instead of an alpha-beta search with domain-specific enhancements, \emph{AlphaZero} uses a general-purpose Monte-Carlo tree search (MCTS) algorithm. 
Each search consists of a series of simulated games of self-play that traverse a tree from root $s_{root}$ to leaf. Each simulation proceeds by selecting in each state $s$ a move $a$ with low visit count, high move probability and high value (averaged over the leaf states of simulations that selected $a$ from $s$) according to the current neural network $f_\theta$. The search returns a vector $\mathbf{\pi}$ representing a probability distribution over moves, either proportionally or greedily with respect to the visit counts at the root state. 

The parameters $\theta$ of the deep neural network in \emph{AlphaZero} are trained by self-play reinforcement learning, starting from randomly initialised parameters $\theta$. Games are played by selecting moves for both players by MCTS, $a_t \sim \pmb{\pi}_t$. At the end of the game, the terminal position $s_T$ is scored according to the rules of the game to compute the game outcome $z$: $-1$ for a loss, $0$ for a draw, and $+1$ for a win. The neural network parameters $\theta$ are updated so as to minimise the error between the predicted outcome $v_t$ and the game outcome $z$, and to maximise the similarity of the policy vector $\mathbf{p}_t$ to the search probabilities $\pmb{\pi}_t$.  Specifically, the parameters $\theta$ are adjusted by gradient descent on a loss function $l$ that sums over mean-squared error and cross-entropy losses respectively, 
\begin{align}
(\mathbf{p}, v) = f_\theta(s), &&
l = (z - v)^2 - \pmb{\pi}^\top \log \mathbf{p} + c ||\theta||^2
\label{eqn:loss}
\end{align}
where $c$ is a parameter controlling the level of $L_2$ weight regularisation.
The updated parameters are used in subsequent games of self-play.

The \emph{AlphaZero} algorithm described in this paper differs from the original \emph{AlphaGo Zero} algorithm in several respects. 

\emph{AlphaGo Zero} estimates and optimises the probability of winning, assuming binary win/loss outcomes. \emph{AlphaZero} instead estimates and optimises the expected outcome, taking account of draws or potentially other outcomes. 

The rules of Go are invariant to rotation and reflection. This fact was exploited in \emph{AlphaGo} and \emph{AlphaGo Zero} in two ways. First, training data was augmented by generating 8 symmetries for each position. Second, during MCTS, board positions were transformed using a randomly selected rotation or reflection before being evaluated by the neural network, so that the Monte-Carlo evaluation is averaged over different biases. The rules of chess and shogi are asymmetric, and in general symmetries cannot be assumed. \emph{AlphaZero} does not augment the training data and does not transform the board position during MCTS.  

In \emph{AlphaGo Zero}, self-play games were generated by the best player from all previous iterations. After each iteration of training, the performance of the new player was measured against the best player; if it won by a margin of $55\%$ then it replaced the best player and self-play games were subsequently generated by this new player. In contrast, \emph{AlphaZero} simply maintains a single neural network that is updated continually, rather than waiting for an iteration to complete. Self-play games are generated by using the latest parameters for this neural network, omitting the evaluation step and the selection of best player. 

\emph{AlphaGo Zero} tuned the hyper-parameter of its search by Bayesian optimisation. In \emph{AlphaZero} we reuse the same hyper-parameters for all games without game-specific tuning. The sole exception is the noise that is added to the prior policy to ensure exploration \cite{Silver17AG0}; this is scaled in proportion to the typical number of legal moves for that game type.

Like \emph{AlphaGo Zero}, the board state is encoded by spatial planes based only on the basic rules for each game. The actions are encoded by either spatial planes or a flat vector, again based only on the basic rules for each game (see Methods). 


We applied the \emph{AlphaZero} algorithm to chess, shogi, and also Go. Unless otherwise specified, the same algorithm settings, network architecture, and hyper-parameters were used for all three games. We trained a separate instance of \emph{AlphaZero} for each game. Training proceeded for 700,000 steps (mini-batches of size 4,096) starting from randomly initialised parameters, using 5,000 first-generation TPUs \cite{jouppi:tpu} to  generate self-play games and 64 second-generation TPUs to train the neural networks.\footnote{The original \emph{AlphaGo Zero} paper used GPUs to train the neural networks.} Further details of the training procedure are provided in the Methods. 

\begin{figure}
\includegraphics[height=0.160\textheight]{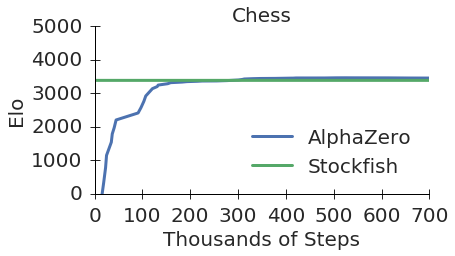}
\includegraphics[height=0.160\textheight]{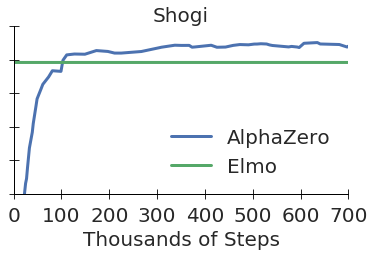}
\includegraphics[height=0.160\textheight]{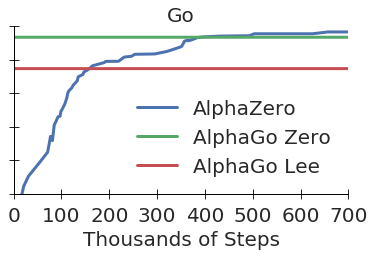}
\caption{
\label{fig:training}
Training \emph{AlphaZero} for 700,000 steps. Elo ratings were computed from evaluation games between different players when given one second per move. 
\textbf{a} Performance of \emph{AlphaZero} in chess, compared to 2016 TCEC world-champion program \emph{Stockfish}.
\textbf{b} Performance of \emph{AlphaZero} in shogi, compared to 2017 CSA world-champion program \emph{Elmo}.
\textbf{c} Performance of \emph{AlphaZero} in Go, compared to \emph{AlphaGo Lee} 
and \emph{AlphaGo Zero} (20 block / 3 day) \cite{Silver17AG0}. 
}
\end{figure}

Figure~\ref{fig:training} shows the performance of \emph{AlphaZero} during self-play reinforcement learning, as a function of training steps, on an Elo scale \cite{coulom:bayeselo}. In chess, \emph{AlphaZero} outperformed \emph{Stockfish} after just 4 hours (300k steps); in shogi, \emph{AlphaZero} outperformed \emph{Elmo} after less than 2 hours (110k steps); and in Go, \emph{AlphaZero} outperformed \emph{AlphaGo Lee} \cite{Silver17AG0} after 8 hours (165k steps).\footnote{\emph{AlphaGo Master} and \emph{AlphaGo Zero} were ultimately trained for 100 times this length of time; we do not reproduce that effort here.}

We evaluated the fully trained instances of \emph{AlphaZero} against \emph{Stockfish}, \emph{Elmo} and the previous version of \emph{AlphaGo Zero} (trained for 3 days) in chess, shogi and Go respectively, playing 100 game matches at tournament time controls of one minute per move. \emph{AlphaZero} and the previous \emph{AlphaGo Zero} used a single machine with 4 TPUs. \emph{Stockfish} and \emph{Elmo} played at their strongest skill level using 64 threads and a hash size of 1GB. \emph{AlphaZero} convincingly defeated all opponents, losing zero games to \emph{Stockfish}
and eight games to \emph{Elmo} (see Supplementary Material for several example games), as well as defeating the previous version of \emph{AlphaGo Zero} (see Table \ref{tab:results}).

\begin{table}
\begin{tabularx}{\textwidth}{XXX | XXX}
\toprule
Game & White & Black & Win & Draw & Loss \\
\midrule
\multirow{2}{*}{Chess} & \emph{AlphaZero} & \emph{Stockfish} & 25 & 25 & 0 \\
& \emph{Stockfish} & \emph{AlphaZero} & 3 & 47 & 0 \\
\midrule
\multirow{2}{*}{Shogi} & \emph{AlphaZero} & \emph{Elmo} & 43 & 2 & 5 \\
& \emph{Elmo} & \emph{AlphaZero} & 47 & 0 & 3 \\
\midrule
\multirow{2}{*}{Go} & \emph{AlphaZero} & \emph{AG0 3-day} & 31 & -- & 19 \\
& \emph{AG0 3-day} & \emph{AlphaZero} & 29 & -- & 21 \\ 
\bottomrule
\end{tabularx}
\caption{
\label{tab:results}
Tournament evaluation of \emph{AlphaZero} in chess, shogi, and Go, as games won, drawn or lost from \emph{AlphaZero}'s perspective, in 100 game matches against \emph{Stockfish}, \emph{Elmo}, and the previously published \emph{AlphaGo Zero} after 3 days of training. Each program was given 1 minute of thinking time per move. }
\end{table}

\begin{figure}
\includegraphics[width=0.49\textwidth]{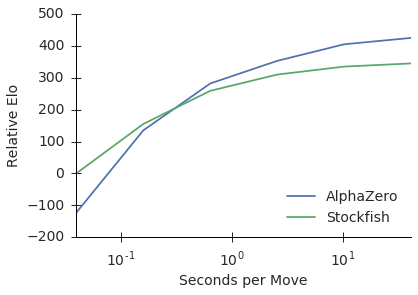}
\includegraphics[width=0.49\textwidth]{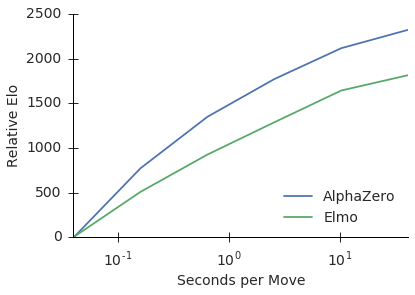}
\caption{
\label{fig:scalability}
Scalability of \emph{AlphaZero} with thinking time, measured on an Elo scale. 
\textbf{a} Performance of \emph{AlphaZero} and \emph{Stockfish} in chess, plotted against thinking time per move. 
\textbf{b} Performance of \emph{AlphaZero} and \emph{Elmo} in shogi, plotted against thinking time per move. 
}
\end{figure}

We also analysed the relative performance of \emph{AlphaZero}'s MCTS search compared to the state-of-the-art alpha-beta search engines used by \emph{Stockfish} and \emph{Elmo}. \emph{AlphaZero} searches just 80 thousand positions per second in chess and 40 thousand in shogi, compared to 70 million for \emph{Stockfish} and 35 million for \emph{Elmo}. \emph{AlphaZero} compensates for the lower number of evaluations by using its deep neural network to focus much more selectively on the most promising variations -- arguably a more ``human-like" approach to search, as originally proposed by Shannon \cite{shannon1950xxii}. Figure~\ref{fig:scalability} shows the scalability of each player with respect to thinking time, measured on an Elo scale, relative to \emph{Stockfish} or \emph{Elmo} with 40ms thinking time. \emph{AlphaZero}'s MCTS scaled more effectively with thinking time than either \emph{Stockfish} or \emph{Elmo}, calling into question the widely held belief \cite{arenz:mcc,david2016deepchess} that alpha-beta search is inherently superior in these domains.\footnote{The prevalence of draws in high-level chess tends to compress the Elo scale, compared to shogi or Go.}

Finally, we analysed the chess knowledge discovered by \emph{AlphaZero}. Table \ref{tab:openings} analyses the most common human openings (those played more than 100,000 times in an online database of human chess games \cite{chess365}). Each of these openings is independently discovered and played frequently by \emph{AlphaZero} during self-play training. When starting from each human opening, \emph{AlphaZero} convincingly defeated \emph{Stockfish}, suggesting that it has indeed mastered a wide spectrum of chess play.

\begin{table}
\scriptsize

\vspace{-10.0em}

\begin{tabularx}{\textwidth}{X c X c}
\toprule

%
%
\multicolumn{2}{c}{A10: English Opening} & \multicolumn{2}{c}{D06: Queen’s Gambit} \\
\newgame \hidemoves{1. c4} \scalebox{0.38}{\showboard} & 
\includegraphics[width=0.3\textwidth]{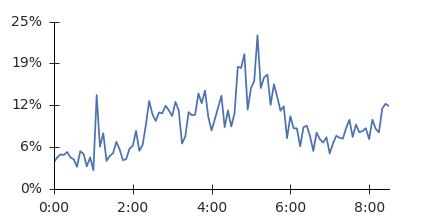} &
\newgame \hidemoves{1. d4 d5 2. c4} \scalebox{0.38}{\showboard} & 
\includegraphics[width=0.3\textwidth]{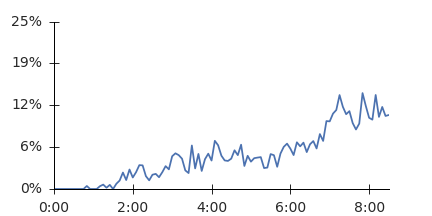} \\
w {\color{darkgreen}20}/{\color{darkgray}30}/{\color{red}0}, b {\color{darkgreen}8}/{\color{darkgray}40}/{\color{red}2} & 1...e5 g3 d5 cxd5 {\symknight}f6 {\symbishop}g2 {\symknight}xd5 {\symknight}f3 & w {\color{darkgreen}16}/{\color{darkgray}34}/{\color{red}0}, b {\color{darkgreen}1}/{\color{darkgray}47}/{\color{red}2} & 2...c6 {\symknight}c3 {\symknight}f6 {\symknight}f3 a6 g3 c4 a4 \\
\midrule

\multicolumn{2}{c}{A46: Queen’s Pawn Game} & \multicolumn{2}{c}{E00: Queen’s Pawn Game} \\
\newgame \hidemoves{1. d4 Nf6 2. Nf3} \scalebox{0.38}{\showboard} & 
\includegraphics[width=0.3\textwidth]{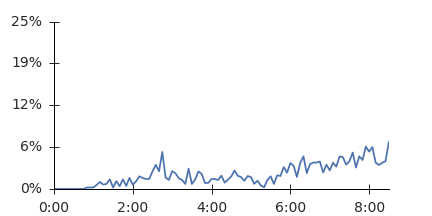} &
\newgame \hidemoves{1. d4 Nf6 2. c4 e6} \scalebox{0.38}{\showboard} & 
\includegraphics[width=0.3\textwidth]{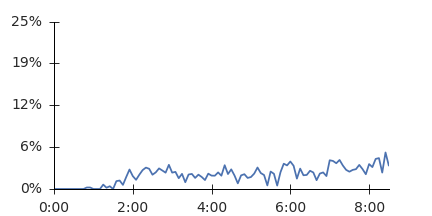} \\
w {\color{darkgreen}24}/{\color{darkgray}26}/{\color{red}0}, b {\color{darkgreen}3}/{\color{darkgray}47}/{\color{red}0} & 2...d5 c4 e6 {\symknight}c3 {\symbishop}e7 {\symbishop}f4 O-O e3 & w {\color{darkgreen}17}/{\color{darkgray}33}/{\color{red}0}, b {\color{darkgreen}5}/{\color{darkgray}44}/{\color{red}1} & 3.{\symknight}f3 d5 {\symknight}c3 {\symbishop}b4 {\symbishop}g5 h6 {\symqueen}a4 {\symknight}c6 \\
\midrule

\multicolumn{2}{c}{E61: King’s Indian Defence} & \multicolumn{2}{c}{C00: French Defence} \\
\newgame \hidemoves{1. d4 Nf6 2. c4 g6 3. Nc3} \scalebox{0.38}{\showboard} & 
\includegraphics[width=0.3\textwidth]{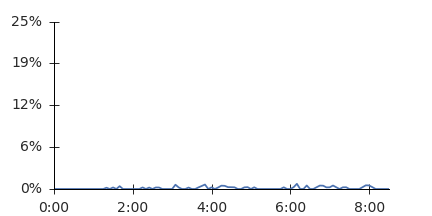} &
\newgame \hidemoves{1. e4 e6 2. d4 d5} \scalebox{0.38}{\showboard} & 
\includegraphics[width=0.3\textwidth]{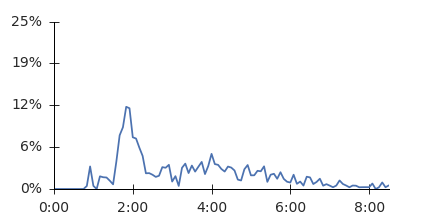} \\
w {\color{darkgreen}16}/{\color{darkgray}34}/{\color{red}0}, b {\color{darkgreen}0}/{\color{darkgray}48}/{\color{red}2} & 3...d5 cxd5 {\symknight}xd5 e4 {\symknight}xc3 bxc3 {\symbishop}g7 {\symbishop}e3 & w {\color{darkgreen}39}/{\color{darkgray}11}/{\color{red}0}, b {\color{darkgreen}4}/{\color{darkgray}46}/{\color{red}0} & 3.{\symknight}c3 {\symknight}f6 e5 {\symknight}d7 f4 c5 {\symknight}f3 {\symbishop}e7 \\
\midrule

\multicolumn{2}{c}{B50: Sicilian Defence} & \multicolumn{2}{c}{B30: Sicilian Defence} \\
\newgame \hidemoves{1. e4 c5 2. Nf3 d6} \scalebox{0.38}{\showboard} & 
\includegraphics[width=0.3\textwidth]{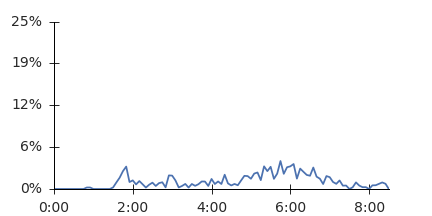} &
\newgame \hidemoves{1. e4 c5 2. Nf3 Nc6} \scalebox{0.38}{\showboard} & 
\includegraphics[width=0.3\textwidth]{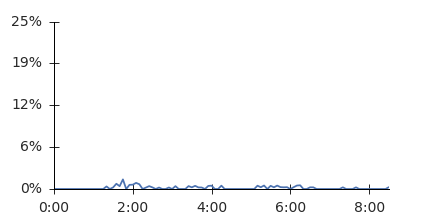} \\
w {\color{darkgreen}17}/{\color{darkgray}32}/{\color{red}1}, b {\color{darkgreen}4}/{\color{darkgray}43}/{\color{red}3} & 3.d4 cxd4 {\symknight}xd4 {\symknight}f6 {\symknight}c3 a6 f3 e5 & w {\color{darkgreen}11}/{\color{darkgray}39}/{\color{red}0}, b {\color{darkgreen}3}/{\color{darkgray}46}/{\color{red}1} & 3.{\symbishop}b5 e6 O-O {\symknight}e7 {\symrook}e1 a6 {\symbishop}f1 d5 \\
\midrule

\multicolumn{2}{c}{B40: Sicilian Defence} & \multicolumn{2}{c}{C60: Ruy Lopez (Spanish Opening)} \\
\newgame \hidemoves{1. e4 c5 2. Nf3 e6} \scalebox{0.38}{\showboard} & 
\includegraphics[width=0.3\textwidth]{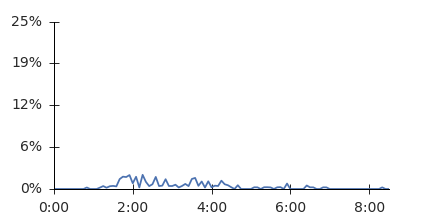} &
\newgame \hidemoves{1. e4 e5 2. Nf3 Nc6 3. Bb5 a6} \scalebox{0.38}{\showboard} & 
\includegraphics[width=0.3\textwidth]{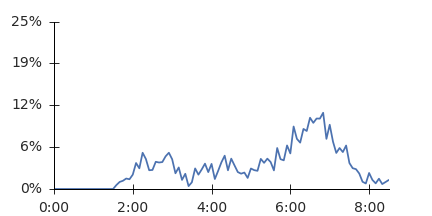} \\
w {\color{darkgreen}17}/{\color{darkgray}31}/{\color{red}2}, b {\color{darkgreen}3}/{\color{darkgray}40}/{\color{red}7} & 3.d4 cxd4 {\symknight}xd4 {\symknight}c6 {\symknight}c3 {\symqueen}c7 {\symbishop}e3 a6 & w {\color{darkgreen}27}/{\color{darkgray}22}/{\color{red}1}, b {\color{darkgreen}6}/{\color{darkgray}44}/{\color{red}0} & 4.{\symbishop}a4 {\symbishop}e7 O-O {\symknight}f6 {\symrook}e1 b5 {\symbishop}b3 O-O \\
\midrule

\multicolumn{2}{c}{B10: Caro-Kann Defence} & \multicolumn{2}{c}{A05: Reti Opening} \\
\newgame \hidemoves{1. e4 c6} \scalebox{0.38}{\showboard} & 
\includegraphics[width=0.3\textwidth]{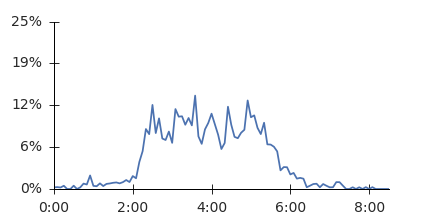} &
\newgame \hidemoves{1. Nf3 Nf6} \scalebox{0.38}{\showboard} & 
\includegraphics[width=0.3\textwidth]{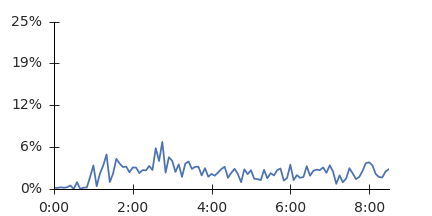} \\
w {\color{darkgreen}25}/{\color{darkgray}25}/{\color{red}0}, b {\color{darkgreen}4}/{\color{darkgray}45}/{\color{red}1} & 2.d4 d5 e5 {\symbishop}f5 {\symknight}f3 e6 {\symbishop}e2 a6 & w {\color{darkgreen}13}/{\color{darkgray}36}/{\color{red}1}, b {\color{darkgreen}7}/{\color{darkgray}43}/{\color{red}0} & 2.c4 e6 d4 d5 {\symknight}c3 {\symbishop}e7 {\symbishop}f4 O-O \\
%
%
\midrule
\multicolumn{2}{c}{Total games: w {\color{darkgreen}242}/{\color{darkgray}353}/{\color{red}5}, b {\color{darkgreen}48}/{\color{darkgray}533}/{\color{red}19}} &
\multicolumn{2}{c}{Overall percentage: w {\color{darkgreen}40.3}/{\color{darkgray}58.8}/{\color{red}0.8}, b {\color{darkgreen}8.0}/{\color{darkgray}88.8}/{\color{red}3.2}} \\
\bottomrule
\end{tabularx}
\caption{
\label{tab:openings}
Analysis of the 12 most popular human openings (played more than 100,000 times in an online database \cite{chess365}). Each opening is labelled by its ECO code and common name. The plot shows the proportion of self-play training games in which \emph{AlphaZero} played each opening, against training time. We also report the win/draw/loss results of 100 game  \emph{AlphaZero} vs. \emph{Stockfish} matches starting from each opening, as either white (w) or black (b), from \emph{AlphaZero}'s perspective. Finally, the principal variation (PV) of \emph{AlphaZero} is provided from each opening. 
}
\end{table}


The game of chess represented the pinnacle of AI research over several decades. State-of-the-art programs are based on powerful engines that search many millions of positions, leveraging handcrafted domain expertise and sophisticated domain adaptations. \emph{AlphaZero} is a generic reinforcement learning algorithm -- originally devised for the game of Go -- that achieved superior results within a few hours, searching a thousand times fewer positions, given no domain knowledge except the rules of chess. Furthermore, the same algorithm was applied without modification to the more challenging game of shogi, again outperforming the state of the art within a few hours. 

\bibliographystyle{plain}
\bibliography{main}

\newpage

\setcounter{table}{0}
\renewcommand{\thetable}{S\arabic{table}}%
\setcounter{figure}{0}
\renewcommand{\thefigure}{S\arabic{figure}}%

\section*{Methods}

\subsection*{Anatomy of a Computer Chess Program}

In this section we describe the components of a typical computer chess program, focusing specifically on \emph{Stockfish}\cite{Stockfish}, an open source program that won the 2016 TCEC computer chess championship. 
For an overview of standard methods, see~\cite{Marsland87}.

Each position $s$ is described by a sparse vector of 
handcrafted features $\phi(s)$, including midgame/endgame-specific material point values, material imbalance tables, piece-square tables, mobility and trapped pieces, pawn structure, king safety, outposts, bishop pair, and other miscellaneous evaluation patterns.
Each feature $\phi_i$ is assigned, by a combination of manual and automatic tuning, a corresponding weight $w_i$ and the position is evaluated by a linear combination $v(s,w) = \phi(s)^\top w$. However, this raw evaluation is only considered accurate for positions that are ``quiet'', with no unresolved captures or checks. A domain-specialised \emph{quiescence search} is used to resolve ongoing tactical situations before the evaluation function is applied.

The final evaluation of a position $s$ is computed by a minimax search that evaluates each leaf using a quiescence search. Alpha-beta pruning is used to safely cut any branch that is provably dominated by another variation. Additional cuts are achieved using aspiration windows and principal variation search.
Other pruning strategies include null move pruning (which assumes a pass move should be worse than any variation, in positions that are unlikely to be in \emph{zugzwang}, as determined by simple heuristics), futility pruning (which assumes knowledge of the maximum possible change in evaluation), and other domain-dependent pruning rules (which assume knowledge of the value of captured pieces).

The search is focused on promising variations both by extending the search depth of promising variations, and by reducing the search depth of unpromising variations based on heuristics like history, static-exchange evaluation (SEE), and moving piece type. Extensions are based on domain-independent rules that identify singular moves with no sensible alternative, and domain-dependent rules, such as extending check moves. Reductions, such as late move reductions, are based heavily on domain knowledge.

The efficiency of alpha-beta search depends critically upon the order in which moves are considered. Moves are therefore ordered by iterative deepening (using a shallower search to order moves for a deeper search). In addition, a combination of domain-independent move ordering heuristics, such as killer heuristic, history heuristic, counter-move heuristic, and also domain-dependent knowledge based on captures (SEE) and potential captures (MVV/LVA).

A transposition table facilitates the reuse of values and move orders when the same position is reached by multiple paths. A carefully tuned opening book is used to select moves at the start of the game. An endgame tablebase, precalculated by exhaustive retrograde analysis of endgame positions, provides the optimal move in all positions with six and sometimes seven pieces or less.

Other strong chess programs, and also earlier programs such as Deep Blue, have used very similar architectures \cite{Marsland87,Campbell02DeepBlue} including the majority of the components described above, although important details vary considerably. 

None of the techniques described in this section are used by \emph{AlphaZero}. It is likely that some of these techniques could further improve the performance of \emph{AlphaZero}; however, we have focused on a pure self-play reinforcement learning approach and leave these extensions for future research.

\subsection*{Prior Work on Computer Chess and Shogi}

In this section we discuss some notable prior work on reinforcement learning in computer chess. 

\emph{NeuroChess} \cite{thrun:neurochess} evaluated positions by a neural network that used 175 handcrafted input features. It was trained by temporal-difference learning to predict the final game outcome, and also the expected features after two moves. \emph{NeuroChess} won 13\% of games against \emph{GnuChess} using a fixed depth 2 search. 

Beal and Smith applied temporal-difference learning to estimate the piece values in chess~\cite{BealSmith00} and shogi~\cite{BealSmith01}, starting from random values and learning solely by self-play.

\emph{KnightCap} \cite{baxter:knightcap} evaluated positions by a neural network that used an attack-table based on knowledge of which squares are attacked or defended by which pieces. It was trained by a variant of temporal-difference learning, known as TD(leaf), that updates the leaf value of the principal variation of an alpha-beta search. \emph{KnightCap} achieved human master level after training against a strong computer opponent with hand-initialised piece-value weights.

\emph{Meep} \cite{veness:rootstrap} evaluated positions by a linear evaluation function based on handcrafted features. It was trained by another variant of temporal-difference learning, known as TreeStrap, that updated all nodes of an alpha-beta search. \emph{Meep} defeated human international master players in 13 out of 15 games, after training by self-play with randomly initialised weights. 

Kaneko and Hoki~\cite{KanekoHoki11} trained the weights of a shogi evaluation function comprising a million features, by learning to select expert human moves during alpha-beta serach. They also performed a large-scale optimization based on minimax search regulated by expert game logs~\cite{HokiKaneko14}; this formed part of the \emph{Bonanza} engine that won the 2013 World Computer Shogi Championship.

\emph{Giraffe} \cite{lai:giraffe} evaluated positions by a neural network that included mobility maps and attack and defend maps describing the lowest valued attacker and defender of each square. It was trained by self-play using TD(leaf), also reaching a standard of play comparable to international masters. 

\emph{DeepChess} \cite{david2016deepchess} trained a neural network to performed pair-wise evaluations of positions. It was trained by supervised learning from a database of human expert games that was pre-filtered to avoid capture moves and drawn games. \emph{DeepChess} reached a strong grandmaster level of play. 

All of these programs combined their learned evaluation functions with an alpha-beta search enhanced by a variety of extensions. 

An approach based on training dual policy and value networks using \emph{AlphaZero}-like policy iteration was successfully applied to improve on the state-of-the-art in Hex \cite{anthony:hex}.

\subsection*{MCTS and Alpha-Beta Search}

For at least four decades the strongest computer chess programs have used alpha-beta search \cite{Marsland87,KnuthMoore75}. \emph{AlphaZero} uses a markedly different approach that averages over the position evaluations within a subtree, rather than computing the minimax evaluation of that subtree. However, chess programs using traditional MCTS were much weaker than alpha-beta search programs, \cite{ramanujan:uct-fail,arenz:mcc}; while alpha-beta programs based on neural networks have previously been unable to compete with faster, handcrafted evaluation functions. 

\emph{AlphaZero} evaluates positions using non-linear function approximation based on a deep neural network, rather than the linear function approximation used in typical chess programs. This provides a much more powerful representation, but may also introduce spurious approximation errors. MCTS averages over these approximation errors, which therefore tend to cancel out when evaluating a large subtree. In contrast, alpha-beta search computes an explicit minimax, which propagates the biggest approximation errors to the root of the subtree. Using MCTS may allow \emph{AlphaZero} to effectively combine its neural network representations with a powerful, domain-independent search.

\subsection*{Domain Knowledge}

\begin{enumerate}
\item The input features describing the position, and the output features describing the move, are structured as a set of planes; i.e. the neural network architecture is matched to the grid-structure of the board.
\item \emph{AlphaZero} is provided with perfect knowledge of the game rules. These are used during MCTS, to simulate the positions resulting from a sequence of moves, to determine game termination, and to score any simulations that reach a terminal state.
\item  Knowledge of the rules is also used to encode the input planes (i.e. castling, repetition, no-progress) and output planes (how pieces move, promotions, and piece drops in shogi).
\item The typical number of legal moves is used to scale the exploration noise (see below).
\item Chess and shogi games exceeding a maximum number of steps (determined by typical game length) were terminated and assigned a drawn outcome; Go games were terminated and scored with Tromp-Taylor rules, similarly to previous work \cite{Silver17AG0}.
\end{enumerate}

\emph{AlphaZero} did not use any form of domain knowledge beyond the points listed above. 

\subsection*{Representation}

In this section we describe the representation of the board inputs, and the representation of the action outputs, used by the neural network in \emph{AlphaZero}. Other representations could have been used; in our experiments the training algorithm worked robustly for many reasonable choices.

The input to the neural network is an $N \times N \times (M T + L)$ image stack that represents state using a concatenation of $T$ sets of $M$ planes of size $N \times N$. Each set of planes represents the board position at a time-step $t - T + 1, ..., t$, and is set to zero for time-steps less than 1. The board is oriented to the perspective of the current player. The $M$ feature planes are composed of binary feature planes indicating the presence of the player's pieces, with one plane for each piece type, and a second set of planes indicating the presence of the opponent's pieces. For shogi there are additional planes indicating the number of captured prisoners of each type. There are an additional $L$  constant-valued input planes denoting the player's colour, the total move count, and the state of special rules: the legality of castling in chess (kingside or queenside); the repetition count for that position (3 repetitions is an automatic draw in chess; 4 in shogi); and the number of moves without progress in chess (50 moves without progress is an automatic draw). Input features are summarised in Table \ref{tab:features}.

\begin{table}
\begin{tabularx}{\textwidth}{Xr | Xr | Xr}
\toprule
\multicolumn{2}{c}{Go} & \multicolumn{2}{|c|}{Chess} & \multicolumn{2}{c}{Shogi} \\
Feature & Planes & Feature & Planes & Feature & Planes \\
\midrule
P1 stone & 1 & P1 piece & 6 & P1 piece & 14 \\
P2 stone & 1 & P2 piece & 6 & P2 piece & 14 \\
&& Repetitions & 2 & Repetitions & 3 \\
&& && P1 prisoner count & 7 \\
&& && P2 prisoner count & 7 \\
\midrule
Colour & 1 & Colour & 1 & Colour & 1 \\
&& Total move count & 1 & Total move count & 1 \\
&& P1 castling & 2 &&\\
&& P2 castling & 2 &&\\
&& No-progress count & 1 &&\\
\midrule
Total & 17 & Total & 119 & Total & 362 \\
\bottomrule
\end{tabularx}
\caption
{
    \label{tab:features}
    Input features used by \emph{AlphaZero} in Go, Chess and Shogi respectively. The first set of features are repeated for each position in a $T=8$-step history. Counts are represented by a single real-valued input; other input features are represented by a one-hot encoding using the specified number of binary input planes. The current player is denoted by P1 and the opponent by P2.
}
\end{table}

\begin{table}
\begin{tabularx}{\textwidth}{Xr | Xr}
\toprule
\multicolumn{2}{c|}{Chess} & \multicolumn{2}{c}{Shogi} \\
Feature & Planes & Feature & Planes \\
\midrule
Queen moves & 56 & Queen moves & 64 \\
Knight moves & 8 & Knight moves & 2 \\
Underpromotions & 9 & Promoting queen moves & 64 \\
&& Promoting knight moves & 2 \\
&& Drop & 7 \\
\midrule
Total & 73 & Total & 139 \\
\bottomrule
\end{tabularx}
\caption
{
    \label{tab:actions}
    Action representation used by \emph{AlphaZero} in Chess and Shogi respectively. The policy is represented by a stack of planes encoding a probability distribution over legal moves; planes correspond to the entries in the table.  
}
\end{table}

A move in chess may be described in two parts: selecting the piece to move, and then selecting among the legal moves for that piece. We represent the policy $\pi(a|s)$ by a $8 \times 8 \times 73$ stack of planes encoding a probability distribution over 4,672 possible moves. Each of the $8 \times 8$ positions identifies the square from which to ``pick up" a piece. The first 56 planes encode possible `queen moves' for any piece: a number of squares $[1..7]$ in which the piece will be moved, along one of eight relative compass directions $\{ N,NE,E,SE,S,SW,W,NW \}$. The next 8 planes encode possible knight moves for that piece. The final 9 planes encode possible underpromotions for pawn moves or captures in two possible diagonals, to knight, bishop or rook respectively. Other pawn moves or captures from the seventh rank are promoted to a queen. 

The policy in shogi is represented by a $9 \times 9 \times 139$ stack of planes similarly encoding a probability distribution over 11,259 possible moves. The first 64 planes encode `queen moves' and the next 2 moves encode knight moves. An additional $64 + 2$ planes encode promoting queen moves and promoting knight moves respectively. The last 7 planes encode a captured piece dropped back into the board at that location. 

The policy in Go is represented identically to \emph{AlphaGo Zero} \cite{Silver17AG0}, using a flat distribution over $19 \times 19 + 1$ moves representing possible stone placements and the pass move. 
We also tried using a flat distribution over moves for chess and shogi; the final result was almost identical although training was slightly slower. 

The action representations are summarised in Table \ref{tab:actions}. Illegal moves are masked out by setting their probabilities to zero, and re-normalising the probabilities for remaining moves.

\subsection*{Configuration}

During training, each MCTS used 800 simulations. The number of games, positions, and thinking time varied per game due largely to different board sizes and game lengths, and are shown in Table \ref{tab:settings}. The learning rate was set to 0.2 for each game, and was dropped three times (to 0.02, 0.002 and 0.0002 respectively) during the course of training. Moves are selected in proportion to the root visit count. Dirichlet noise $\text{Dir}(\alpha)$ was added to the prior probabilities in the root node; this was scaled in inverse proportion to the approximate number of legal moves in a typical position, to a value of $\alpha = \{0.3, 0.15, 0.03\}$ for chess, shogi and Go respectively. Unless otherwise specified, the training and search algorithm and parameters are identical to \emph{AlphaGo Zero}~\cite{Silver17AG0}.

During evaluation, \emph{AlphaZero} selects moves greedily with respect to the root visit count. Each MCTS was executed on a single machine with 4 TPUs.

\begin{table}
\begin{tabularx}{\textwidth}{XXXX}
\toprule
& Chess & Shogi & Go \\
\midrule
Mini-batches & 700k & 700k & 700k \\
Training Time & 9h & 12h & 34h \\
Training Games & 44 million & 24 million & 21 million \\
Thinking Time & 800 sims & 800 sims & 800 sims  \\
              & ~40 ms & ~80 ms & ~200 ms \\
\bottomrule
\end{tabularx}
\caption
{
\label{tab:settings}
Selected statistics of \emph{AlphaZero} training in Chess, Shogi and Go. 
}
\end{table}

\subsection*{Evaluation}

To evaluate performance in chess, we used \emph{Stockfish} version 8 (official Linux release) as a baseline program, using 64 CPU threads and a hash size of 1GB.

To evaluate performance in shogi, we used \emph{Elmo} version WCSC27 in combination with YaneuraOu 2017 Early KPPT 4.73 64AVX2 with 64 CPU threads and a hash size of 1GB with the usi option of EnteringKingRule set to NoEnteringKing.

We evaluated the relative strength of \emph{AlphaZero} (Figure \ref{fig:training}) by measuring the Elo rating of each player. We estimate the probability that player $a$ will defeat player $b$ by a logistic function $p(a \text{ defeats } b) = \frac{1}{1 + \exp{(c_{\mathrm{elo}} (e(b) - e(a))}}$, and estimate the ratings $e(\cdot)$ by Bayesian logistic regression, computed by the \emph{BayesElo} program~\cite{coulom:bayeselo} using the standard constant $c_{\mathrm{elo}} = 1/400$. 
Elo ratings were computed from the results of a 1 second per move tournament between iterations of \emph{AlphaZero} during training, and also a baseline player: either \emph{Stockfish}, \emph{Elmo} or \emph{AlphaGo Lee} respectively. The Elo rating of the baseline players was anchored to publicly available values \cite{Silver17AG0}. 

We also measured the head-to-head performance of \emph{AlphaZero} against each baseline player. Settings were chosen to correspond with computer chess tournament conditions: each player was allowed 1 minute per move, resignation was enabled for all players (-900 centipawns for 10 consecutive moves for \emph{Stockfish} and \emph{Elmo}, 5\% winrate for \emph{AlphaZero}). Pondering was disabled for all players.

\begin{table}
\begin{tabularx}{\textwidth}{XXXX}
\toprule
Program & Chess & Shogi & Go \\
\midrule
\emph{AlphaZero} & 80k & 40k & 16k \\
\emph{Stockfish} & 70,000k & \\
\emph{Elmo} && 35,000k & \\
\bottomrule
\end{tabularx}
\caption{
\label{tab:speed}
Evaluation speed (positions/second) of \emph{AlphaZero}, \emph{Stockfish}, and \emph{Elmo} in chess, shogi and Go.
}
\end{table}

\subsection*{Example games}

In this section we include 10 example games played by \emph{AlphaZero} against \emph{Stockfish} during the 100 game match using 1 minute per move.

\begin{small}
\begin{longtable}{p{\textwidth}}
\hline
White: \emph{Stockfish} Black: \emph{AlphaZero} \\\nopagebreak
1. e4 e5 2. Nf3 Nc6 3. Bb5 Nf6 4. d3 Bc5 5. Bxc6 dxc6 6. 0-0 Nd7 7. Nbd2 0-0 8. Qe1 f6 9. Nc4 Rf7 10. a4 Bf8 11. Kh1 Nc5 12. a5 Ne6 13. Ncxe5 fxe5 14. Nxe5 Rf6 15. Ng4 Rf7 16. Ne5 Re7 17. a6 c5 18. f4 Qe8 19. axb7 Bxb7 20. Qa5 Nd4 21. Qc3 Re6 22. Be3 Rb6 23. Nc4 Rb4 24. b3 a5 25. Rxa5 Rxa5 26. Nxa5 Ba6 27. Bxd4 Rxd4 28. Nc4 Rd8 29. g3 h6 30. Qa5 Bc8 31. Qxc7 Bh3 32. Rg1 Rd7 33. Qe5 Qxe5 34. Nxe5 Ra7 35. Nc4 g5 36. Rc1 Bg7 37. Ne5 Ra8 38. Nf3 Bb2 39. Rb1 Bc3 40. Ng1 Bd7 41. Ne2 Bd2 42. Rd1 Be3 43. Kg2 Bg4 44. Re1 Bd2 45. Rf1 Ra2 46. h3 Bxe2 47. Rf2 Bxf4 48. Rxe2 Be5 49. Rf2 Kg7 50. g4 Bd4 51. Re2 Kf6 52. e5+ Bxe5 53. Kf3 Ra1 54. Rf2 Re1 55. Kg2+ Bf4 56. c3 Rc1 57. d4 Rxc3 58. dxc5 Rxc5 59. b4 Rc3 60. h4 Ke5 61. hxg5 hxg5 62. Re2+ Kf6 63. Kf2 Be5 64. Ra2 Rc4 65. Ra6+ Ke7 66. Ra5 Ke6 67. Ra6+ Bd6 0-1 \\
\hline
White: \emph{Stockfish} Black: \emph{AlphaZero} \\\nopagebreak
1. e4 e5 2. Nf3 Nc6 3. Bb5 Nf6 4. d3 Bc5 5. Bxc6 dxc6 6. 0-0 Nd7 7. c3 0-0 8. d4 Bd6 9. Bg5 Qe8 10. Re1 f6 11. Bh4 Qf7 12. Nbd2 a5 13. Bg3 Re8 14. Qc2 Nf8 15. c4 c5 16. d5 b6 17. Nh4 g6 18. Nhf3 Bd7 19. Rad1 Re7 20. h3 Qg7 21. Qc3 Rae8 22. a3 h6 23. Bh4 Rf7 24. Bg3 Rfe7 25. Bh4 Rf7 26. Bg3 a4 27. Kh1 Rfe7 28. Bh4 Rf7 29. Bg3 Rfe7 30. Bh4 g5 31. Bg3 Ng6 32. Nf1 Rf7 33. Ne3 Ne7 34. Qd3 h5 35. h4 Nc8 36. Re2 g4 37. Nd2 Qh7 38. Kg1 Bf8 39. Nb1 Nd6 40. Nc3 Bh6 41. Rf1 Ra8 42. Kh2 Kf8 43. Kg1 Qg6 44. f4 gxf3 45. Rxf3 Bxe3+ 46. Rfxe3 Ke7 47. Be1 Qh7 48. Rg3 Rg7 49. Rxg7+ Qxg7 50. Re3 Rg8 51. Rg3 Qh8 52. Nb1 Rxg3 53. Bxg3 Qh6 54. Nd2 Bg4 55. Kh2 Kd7 56. b3 axb3 57. Nxb3 Qg6 58. Nd2 Bd1 59. Nf3 Ba4 60. Nd2 Ke7 61. Bf2 Qg4 62. Qf3 Bd1 63. Qxg4 Bxg4 64. a4 Nb7 65. Nb1 Na5 66. Be3 Nxc4 67. Bc1 Bd7 68. Nc3 c6 69. Kg1 cxd5 70. exd5 Bf5 71. Kf2 Nd6 72. Be3 Ne4+ 73. Nxe4 Bxe4 74. a5 bxa5 75. Bxc5+ Kd7 76. d6 Bf5 77. Ba3 Kc6 78. Ke1 Kd5 79. Kd2 Ke4 80. Bb2 Kf4 81. Bc1 Kg3 82. Ke2 a4 83. Kf1 Kxh4 84. Kf2 Kg4 85. Ba3 Bd7 86. Bc1 Kf5 87. Ke3 Ke6 0-1 \\
\hline
White: \emph{AlphaZero} Black: \emph{Stockfish} \\\nopagebreak
1. Nf3 Nf6 2. c4 b6 3. d4 e6 4. g3 Ba6 5. Qc2 c5 6. d5 exd5 7. cxd5 Bb7 8. Bg2 Nxd5 9. 0-0 Nc6 10. Rd1 Be7 11. Qf5 Nf6 12. e4 g6 13. Qf4 0-0 14. e5 Nh5 15. Qg4 Re8 16. Nc3 Qb8 17. Nd5 Bf8 18. Bf4 Qc8 19. h3 Ne7 20. Ne3 Bc6 21. Rd6 Ng7 22. Rf6 Qb7 23. Bh6 Nd5 24. Nxd5 Bxd5 25. Rd1 Ne6 26. Bxf8 Rxf8 27. Qh4 Bc6 28. Qh6 Rae8 29. Rd6 Bxf3 30. Bxf3 Qa6 31. h4 Qa5 32. Rd1 c4 33. Rd5 Qe1+ 34. Kg2 c3 35. bxc3 Qxc3 36. h5 Re7 37. Bd1 Qe1 38. Bb3 Rd8 39. Rf3 Qe4 40. Qd2 Qg4 41. Bd1 Qe4 42. h6 Nc7 43. Rd6 Ne6 44. Bb3 Qxe5 45. Rd5 Qh8 46. Qb4 Nc5 47. Rxc5 bxc5 48. Qh4 Rde8 49. Rf6 Rf8 50. Qf4 a5 51. g4 d5 52. Bxd5 Rd7 53. Bc4 a4 54. g5 a3 55. Qf3 Rc7 56. Qxa3 Qxf6 57. gxf6 Rfc8 58. Qd3 Rf8 59. Qd6 Rfc8 60. a4 1-0 \\
\hline
White: \emph{AlphaZero} Black: \emph{Stockfish} \\\nopagebreak
1. d4 e6 2. Nc3 Nf6 3. e4 d5 4. e5 Nfd7 5. f4 c5 6. Nf3 Nc6 7. Be3 Be7 8. Qd2 a6 9. Bd3 c4 10. Be2 b5 11. a3 Rb8 12. 0-0 0-0 13. f5 a5 14. fxe6 fxe6 15. Bd1 b4 16. axb4 axb4 17. Ne2 c3 18. bxc3 Nb6 19. Qe1 Nc4 20. Bc1 bxc3 21. Qxc3 Qb6 22. Kh1 Nb2 23. Nf4 Nxd1 24. Rxd1 Bd7 25. h4 Ra8 26. Bd2 Rfb8 27. h5 Rxa1 28. Rxa1 Qb2 29. Qxb2 Rxb2 30. c3 Rb3 31. Ra8+ Rb8 32. Ra2 Rb3 33. g4 Ra3 34. Rb2 Kf7 35. Kg2 Bc8 36. Rb6 Ra6 37. Rb1 Ke8 38. Kg3 h6 39. Ng6 Ra3 40. Rb6 Bd7 41. g5 hxg5 42. Kg4 Bd8 43. Rb2 Bc8 44. Nxg5 Ra1 45. Nf3 Ra3 46. Be1 Ba5 47. Rf2 Ra1 48. Bd2 Bd8 49. Rh2 Ne7 50. Bg5 Nf5 51. Bxd8 Kxd8 52. Rb2 Rc1 53. Ngh4 Nxh4 54. Nxh4 Bd7 55. Rb8+ Bc8 56. Ng2 Rxc3 57. Nf4 Rc1 58. Ra8 Kd7 59. Kf3 Rc3+ 60. Kf2 Ke7 61. Kg2 Kf7 62. Ng6 Ke8 63. Ra1 Rc7 64. Kh3 Rf7 65. Kg4 Kd8 66. Nf4 Bd7 67. Ra7 Kc8 68. Kg3 Re7 69. Nd3 Kb8 70. Ra6 Bc8 71. Rb6+ Kc7 72. Rd6 Kb8 73. Nc5 g6 74. h6 Rh7 75. Nxe6 Rxh6 76. Nf4 Rh1 77. Nxd5 Rh3+ 78. Kf4 Rh4+ 79. Ke3 Rh3+ 80. Kd2 Bf5 81. Ne7 Rh2+ 82. Ke3 Bh3 83. Nxg6 Rh1 84. Nf4 Bg4 85. Rf6 Kc7 86. Nd3 Bd7 87. d5 Bb5 88. Nf4 Ba4 89. Kd4 Be8 90. Rf8 Rd1+ 91. Kc5 Rc1+ 92. Kb4 Rb1+ 93. Kc3 Bb5 94. Kd4 Ba6 95. Rf7+ 1-0 \\
\hline
White: \emph{AlphaZero} Black: \emph{Stockfish} \\\nopagebreak
1. d4 Nf6 2. c4 e6 3. Nf3 b6 4. g3 Bb7 5. Bg2 Be7 6. 0-0 0-0 7. d5 exd5 8. Nh4 c6 9. cxd5 Nxd5 10. Nf5 Nc7 11. e4 Bf6 12. Nd6 Ba6 13. Re1 Ne8 14. e5 Nxd6 15. exf6 Qxf6 16. Nc3 Nb7 17. Ne4 Qg6 18. h4 h6 19. h5 Qh7 20. Qg4 Kh8 21. Bg5 f5 22. Qf4 Nc5 23. Be7 Nd3 24. Qd6 Nxe1 25. Rxe1 fxe4 26. Bxe4 Rf5 27. Bh4 Bc4 28. g4 Rd5 29. Bxd5 Bxd5 30. Re8+ Bg8 31. Bg3 c5 32. Qd5 d6 33. Qxa8 Nd7 34. Qe4 Nf6 35. Qxh7+ Kxh7 36. Re7 Nxg4 37. Rxa7 Nf6 38. Bxd6 Be6 39. Be5 Nd7 40. Bc3 g6 41. Bd2 gxh5 42. a3 Kg6 43. Bf4 Kf5 44. Bc7 h4 45. Ra8 h5 46. Rh8 Kg6 47. Rd8 Kf7 48. f3 Bf5 49. Bh2 h3 50. Rh8 Kg6 51. Re8 Kf7 52. Re1 Be6 53. Bc7 b5 54. Kh2 Kf6 55. Re3 Ke7 56. Re4 Kf7 57. Bd6 Kf6 58. Kg3 Kf7 59. Kf2 Bf5 60. Re1 Kg6 61. Kg1 c4 62. Kh2 h4 63. Be7 Nb6 64. Bxh4 Na4 65. Re2 Nc5 66. Re5 Nb3 67. Rd5 Be6 68. Rd6 Kf5 69. Be1 Ke5 70. Rb6 Bd7 71. Kg3 Nc1 72. Rh6 Kd5 73. Bc3 Bf5 74. Rh5 Ke6 75. Kf2 Nd3+ 76. Kg1 Nf4 77. Rh6+ Ke7 78. Kh2 Nd5 79. Kg3 Be6 80. Rh5 Ke8 81. Re5 Kf7 82. Bd2 Ne7 83. Bb4 Nd5 84. Bc3 Ke7 85. Bd2 Kf6 86. f4 Ne7 87. Rxb5 Nf5+ 88. Kh2 Ke7 89. Ra5 Nh4 90. Bb4+ Kf7 91. Rh5 Nf3+ 92. Kg3 Kg6 93. Rh8 Nd4 94. Bc3 Nf5+ 95. Kxh3 Bd7 96. Kh2 Kf7 97. Rb8 Ke6 98. Kg1 Bc6 99. Rb6 Kd5 100. Kf2 Bd7 101. Ke1 Ke4 102. Bd2 Kd5 103. Rf6 Nd6 104. Rh6 Nf5 105. Rh8 Ke4 106. Rh7 Bc8 107. Rc7 Ba6 108. Rc6 Bb5 109. Rc5 Bd7 110. Rxc4+ Kd5 111. Rc7 Kd6 112. Rc3 Ke6 113. Rc5 Nd4 114. Be3 Nf5 115. Bf2 Nd6 116. Rc3 Ne4 117. Rd3 1-0 \\
\hline
White: \emph{AlphaZero} Black: \emph{Stockfish} \\\nopagebreak
1. d4 Nf6 2. Nf3 e6 3. c4 b6 4. g3 Be7 5. Bg2 Bb7 6. 0-0 0-0 7. d5 exd5 8. Nh4 c6 9. cxd5 Nxd5 10. Nf5 Nc7 11. e4 Bf6 12. Nd6 Ba6 13. Re1 Ne8 14. e5 Nxd6 15. exf6 Qxf6 16. Nc3 Bc4 17. h4 h6 18. b3 Qxc3 19. Bf4 Nb7 20. bxc4 Qf6 21. Be4 Na6 22. Be5 Qe6 23. Bd3 f6 24. Bd4 Qf7 25. Qg4 Rfd8 26. Re3 Nac5 27. Bg6 Qf8 28. Rd1 Rab8 29. Kg2 Ne6 30. Bc3 Nbc5 31. Rde1 Na4 32. Bd2 Kh8 33. f4 Qd6 34. Bc1 Nd4 35. Re7 f5 36. Bxf5 Nxf5 37. Qxf5 Rf8 38. Rxd7 Rxf5 39. Rxd6 Rf7 40. g4 Kg8 41. g5 hxg5 42. hxg5 Nc5 43. Kf3 Nb7 44. Rdd1 Na5 45. Re4 c5 46. Bb2 Nc6 47. g6 Rc7 48. Kg4 Nd4 49. Rd2 Rf8 50. Bxd4 cxd4 51. Rdxd4 Rfc8 52. Kg5 Rf8 53. Rd2 Rc6 54. Rd5 Rc7 55. f5 Rb7 56. a3 Rc7 57. a4 a6 58. Red4 Rcc8 59. Re5 Rc7 60. a5 Rc5 61. Rxc5 bxc5 62. Rd6 Ra8 63. Re6 Kf8 64. Rc6 Ke7 65. Kf4 Kd7 66. Rxc5 Rh8 67. Rd5+ Ke7 68. Re5+ Kd7 69. Re6 Rh4+ 70. Kg5 1-0 \\
\hline
White: \emph{AlphaZero} Black: \emph{Stockfish} \\\nopagebreak
1. d4 Nf6 2. c4 e6 3. Nf3 b6 4. g3 Bb7 5. Bg2 Bb4+ 6. Bd2 Bxd2+ 7. Qxd2 d5 8. 0-0 0-0 9. cxd5 exd5 10. Nc3 Nbd7 11. b4 c6 12. Qb2 a5 13. b5 c5 14. Rac1 Qe7 15. Na4 Rab8 16. Rfd1 c4 17. Ne5 Qe6 18. f4 Rfd8 19. Qd2 Nf8 20. Nc3 Ng6 21. Rf1 Qd6 22. a4 Rbc8 23. e3 Ne7 24. g4 Ne8 25. f5 f6 26. Nf3 Qd7 27. Qf2 Nd6 28. Nd2 Rf8 29. Qg3 Rcd8 30. Rf4 Nf7 31. Rf2 Rfe8 32. h3 Qd6 33. Nf1 Qa3 34. Rcc2 h5 35. Qc7 Qd6 36. Qxd6 Rxd6 37. Ng3 h4 38. Nh5 Ng5 39. Rf1 Kh7 40. Nf4 Rdd8 41. Kh2 Rd7 42. Bh1 Rd6 43. Ng2 g6 44. Nxh4 gxf5 45. gxf5 Rh8 46. Nf3 Kg7 47. Nxg5 fxg5 48. Rg2 Kf6 49. Rg3 Re8 50. Bf3 Rdd8 51. Be2 Rf8 52. Bg4 Nc8 53. Bf3 Rfe8 54. h4 Rh8 55. h5 Rhe8 56. Bg2 Ne7 57. h6 Rh8 58. Rh3 Rh7 59. Kg1 Ba8 60. Nd1 g4 61. Rh5 g3 62. Nc3 Ng8 63. Ne2 Rxh6 64. Nxg3 Rxh5 65. Nxh5+ Kf7 66. Kf2 Nf6 67. Nxf6 Kxf6 68. Rh1 c3 69. Rc1 Rh8 70. Rxc3 Kxf5 71. Rc7 Kf6 72. Bf3 Rg8 73. Rh7 Rg6 74. Bd1 Rg8 75. Rh6+ Ke7 76. Rxb6 Kd7 77. Rf6 Ke7 78. Rh6 Rg7 79. Rh8 Bb7 80. Rh5 Kd6 81. Rh3 Rf7+ 82. Ke1 Bc8 83. Rh6+ Kc7 84. Rc6+ Kb8 85. Rd6 Bb7 86. b6 Ba6 87. Rxd5 Rf6 88. Rxa5 Rxb6 89. Kd2 Bb7 90. Rb5 Rf6 91. Bb3 Kc7 92. Re5 Ba6 93. Kc3 Rf1 94. Bc2 Rh1 95. a5 Kd6 96. e4 Bf1 97. Rf5 Bg2 98. Rf4 Rc1 99. Kb2 Rh1 100. a6 1-0 \\
\hline
White: \emph{AlphaZero} Black: \emph{Stockfish} \\\nopagebreak
1. d4 Nf6 2. c4 e6 3. Nf3 b6 4. g3 Bb7 5. Bg2 Bb4+ 6. Bd2 Be7 7. Nc3 c6 8. e4 d5 9. e5 Ne4 10. 0-0 Ba6 11. b3 Nxc3 12. Bxc3 dxc4 13. b4 b5 14. Nd2 0-0 15. Ne4 Bb7 16. Qg4 Nd7 17. Nc5 Nxc5 18. dxc5 a5 19. a3 axb4 20. axb4 Rxa1 21. Rxa1 Qd3 22. Rc1 Ra8 23. h4 Qd8 24. Be4 Qc8 25. Kg2 Qc7 26. Qh5 g6 27. Qg4 Bf8 28. h5 Rd8 29. Qh4 Qe7 30. Qf6 Qe8 31. Rh1 Rd7 32. hxg6 fxg6 33. Qh4 Qe7 34. Qg4 Rd8 35. Bb2 Qf7 36. Bc1 c3 37. Be3 Be7 38. Qe2 Bf8 39. Qc2 Bg7 40. Qxc3 Qd7 41. Rc1 Qc7 42. Bg5 Rf8 43. f4 h6 44. Bf6 Bxf6 45. exf6 Qf7 46. Ra1 Qxf6 47. Qxf6 Rxf6 48. Ra7 Rf7 49. Bxg6 Rd7 50. Kf2 Kf8 51. g4 Bc8 52. Ra8 Rc7 53. Ke3 h5 54. gxh5 Kg7 55. Ra2 Re7 56. Be4 e5 57. Bxc6 exf4+ 58. Kxf4 Rf7+ 59. Ke5 Rf5+ 60. Kd6 Rxh5 61. Rg2+ Kf6 62. Kc7 Bf5 63. Kb6 Rh4 64. Ka5 Bg4 65. Bxb5 Ke7 66. Rg3 Bc8 67. Re3+ Kf7 68. Be2 1-0 \\
\hline
White: \emph{AlphaZero}, Black: \emph{Stockfish} \\\nopagebreak
1. d4 e6 2. e4 d5 3. Nc3 Nf6 4. e5 Nfd7 5. f4 c5 6. Nf3 cxd4 7. Nb5 Bb4+ 8. Bd2 Bc5 9. b4 Be7 10. Nbxd4 Nc6 11. c3 a5 12. b5 Nxd4 13. cxd4 Nb6 14. a4 Nc4 15. Bd3 Nxd2 16. Kxd2 Bd7 17. Ke3 b6 18. g4 h5 19. Qg1 hxg4 20. Qxg4 Bf8 21. h4 Qe7 22. Rhc1 g6 23. Rc2 Kd8 24. Rac1 Qe8 25. Rc7 Rc8 26. Rxc8+ Bxc8 27. Rc6 Bb7 28. Rc2 Kd7 29. Ng5 Be7 30. Bxg6 Bxg5 31. Qxg5 fxg6 32. f5 Rg8 33. Qh6 Qf7 34. f6 Kd8 35. Kd2 Kd7 36. Rc1 Kd8 37. Qe3 Qf8 38. Qc3 Qb4 39. Qxb4 axb4 40. Rg1 b3 41. Kc3 Bc8 42. Kxb3 Bd7 43. Kb4 Be8 44. Ra1 Kc7 45. a5 Bd7 46. axb6+ Kxb6 47. Ra6+ Kb7 48. Kc5 Rd8 49. Ra2 Rc8+ 50. Kd6 Be8 51. Ke7 g5 52. hxg5 1-0 \\
\hline
White: \emph{AlphaZero}, Black: \emph{Stockfish} \\\nopagebreak
1. Nf3 Nf6 2. d4 e6 3. c4 b6 4. g3 Bb7 5. Bg2 Be7 6. 0-0 0-0 7. d5 exd5 8. Nh4 c6 9. cxd5 Nxd5 10. Nf5 Nc7 11. e4 d5 12. exd5 Nxd5 13. Nc3 Nxc3 14. Qg4 g6 15. Nh6+ Kg7 16. bxc3 Bc8 17. Qf4 Qd6 18. Qa4 g5 19. Re1 Kxh6 20. h4 f6 21. Be3 Bf5 22. Rad1 Qa3 23. Qc4 b5 24. hxg5+ fxg5 25. Qh4+ Kg6 26. Qh1 Kg7 27. Be4 Bg6 28. Bxg6 hxg6 29. Qh3 Bf6 30. Kg2 Qxa2 31. Rh1 Qg8 32. c4 Re8 33. Bd4 Bxd4 34. Rxd4 Rd8 35. Rxd8 Qxd8 36. Qe6 Nd7 37. Rd1 Nc5 38. Rxd8 Nxe6 39. Rxa8 Kf6 40. cxb5 cxb5 41. Kf3 Nd4+ 42. Ke4 Nc6 43. Rc8 Ne7 44. Rb8 Nf5 45. g4 Nh6 46. f3 Nf7 47. Ra8 Nd6+ 48. Kd5 Nc4 49. Rxa7 Ne3+ 50. Ke4 Nc4 51. Ra6+ Kg7 52. Rc6 Kf7 53. Rc5 Ke6 54. Rxg5 Kf6 55. Rc5 g5 56. Kd4 1-0 \\
\hline
\end{longtable}
\end{small}

\end{document}